\documentclass[10pt,twocolumn,letterpaper]{article}
\usepackage[utf8]{inputenc}
\usepackage{graphicx}
\usepackage{amsmath}
\usepackage{amssymb}
\usepackage{float}
\usepackage{caption}
\usepackage{url}
\usepackage{hyperref}

\title{Recursive ArUco Markers: A Scalable Fiducial Marker Design for Unmanned Aerial Vehicle Landing Pads}

\author{
Rafael Mu\~noz-Salinas$^{1,*}$, Francisco J. Romero-Ramirez$^2$, Sergio Garrido-Jurado$^1$ \\
$^1$Department of Computer Science and A.I., Universidad de C\'ordoba, 14071, C\'ordoba, Spain \\
$^2$Department of Electronics and Computer Engineering, Universidad de C\'ordoba, 14071, C\'ordoba, Spain \\
$^*$ \texttt{in1musar@uco.es}
}

\date{}

\begin{document}

\maketitle

\begin{abstract}
Unmanned Aerial Vehicles (UAVs) increasingly rely on visual fiducial markers for autonomous navigation and precision landing. However, standard markers suffer from limited operational ranges, becoming undetectable when the camera is either too far or too close. While recursive and fractal markers have been proposed to address this issue, existing approaches either require the marker's center to remain visible, making them vulnerable to occlusion, or are limited in their recursion depth and placement. We propose a novel Recursive ArUco marker design. Our method allows any standard fiducial marker to be transformed into a recursive marker with an arbitrary depth. By employing a modified bit-sampling strategy during detection, we embed complete markers within both the black and white bits of the parent marker. This approach guarantees unlimited recursion depth and robust detection even with partial occlusion, as it does not rely on the marker's center being visible. Furthermore, by maintaining a single, unique identifier across all recursive scales, our proposal provides an extensive dictionary of multiple unique landing pads. This capability allows fleets of UAVs to operate simultaneously, with each drone landing at its designated location ---a feature not supported by existing Fractal and Harco markers due to their structural and dictionary constraints.
\end{abstract}

\section{Introduction}

Visual fiducial markers, such as ArUco \cite{arucopaper,ROMERORAMIREZ201838, GarridoJurado2015,aruco_nano_paper}, AprilTag \cite{wang2016iros}, and ARTag \cite{fiala2005artag}, are ubiquitous in computer vision applications, particularly in robotics and Unmanned Aerial Vehicle (UAV) navigation \cite{semerikov2025vision}. In the context of autonomous UAV landing, visual markers are frequently employed to assist in precise pose estimation \cite{marut_aruco,lebedev2020accurate,wubben2019accurate}. However, a challenge in this application is that the camera-to-marker distance varies significantly during the descent. A single standard marker large enough to be detected at high altitudes will eventually fall outside the camera's field of view at lower altitudes, leading to tracking failure. 

\begin{figure}[H]
\centering
\includegraphics[width=0.9\linewidth]{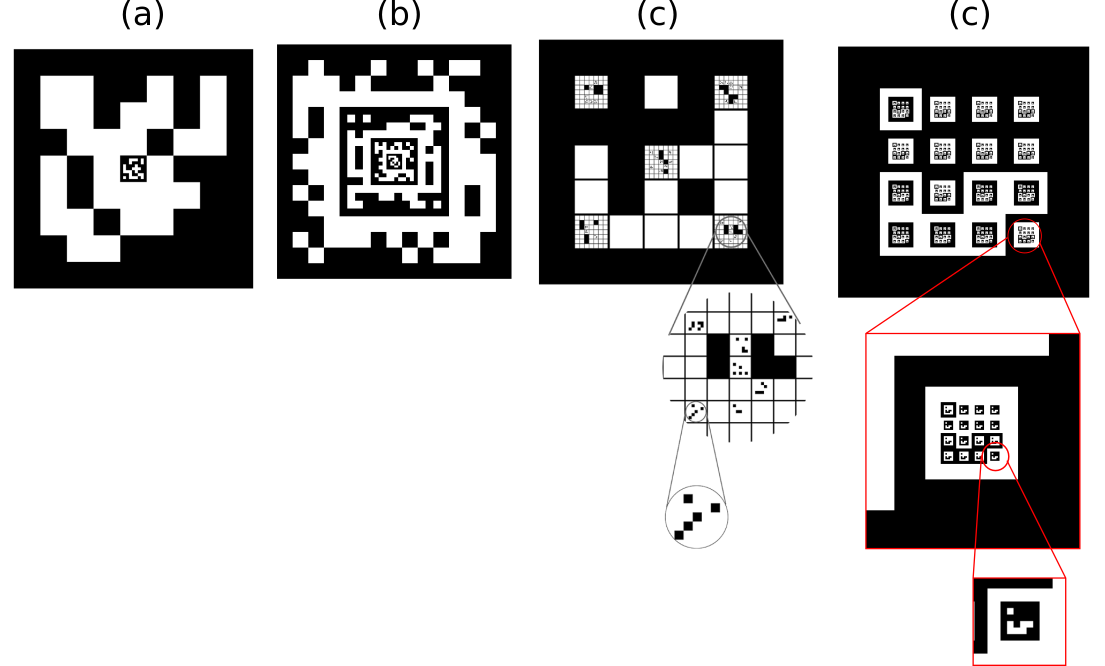}
\caption{Visual comparison of multi-scale markers: (a) Embedded ArUco \cite{khazetdinov2021embedded}. (b) Fractal marker \cite{Fractal}, (c) Harco marker \cite{Harco}, and (d) Our proposed Recursive ArUco marker (RArUco). }
\label{fig:visual_comparison}
\end{figure}

To mitigate this limitation, recursive, nested, or height-adaptive markers have been introduced to provide continuous pose estimation across multiple scales \cite{liu2020vision,wang2023autonomous}. For example, Embedded ArUco \cite{khazetdinov2021embedded} extends the detection range by placing a smaller marker inside the parent marker (see Figure ~\ref{fig:visual_comparison}(a)). Similarly, Fractal markers \cite{Fractal} embed multiple smaller markers within the parent marker (see Figure ~\ref{fig:visual_comparison}(b)), and have been actively applied to UAV landing scenarios \cite{anikin2023autonomous}. However, a critical limitation of both Embedded ArUco and Fractal markers is their dependence on the marker's central region for proper identification; if the center is occluded or falls outside the field of view, the entire hierarchy becomes undetectable. Furthermore, dedicating the center to sub-markers reduces the area available for encoding data. Other approaches, such as Harco markers \cite{Harco}, embed smaller markers strictly within the isolated white pixels of the parent marker to ensure smooth pose transitions (Figure ~\ref{fig:visual_comparison}(c)). However, this approach is limited to a small number of recursion levels, as adding deeper layers inevitably introduces black pixels into the parent's white cells, thereby destroying the upper-level detectability. The placement is also heavily constrained by the parent's binary code. Furthermore, Harco requires separate dictionaries for parent and child markers, meaning the marker's identifier varies with the observed scale, complicating its use in multi-UAV operations.

In this work, we propose the Recursive ArUco marker (Figure \ref{fig:visual_comparison}(d)). Our contribution is a mathematically simple yet highly effective generation and detection scheme that can be applied to any existing dictionary of squared fiducial markers. Unlike Harco markers, our method allows for an arbitrary number of recursion levels and embeds markers within both black and white bits. Crucially, unlike Embedded ArUco and Fractal markers, our detection algorithm does not require the marker's central region to be visible, making it highly robust to occlusions. 

A key aspect of our implementation, specifically designed for drone landing pads, is that we embed the \textit{same} marker into its own bits across all recursive levels. This means that a Recursive ArUco marker maintains a single, unique ID at any scale. Since each drone can be assigned a specific landing pad ID, the drone can unambiguously identify where to land regardless of which part of the marker or which recursive level it sees during its descent. Consequently, a marker dictionary provides as many distinct landing pads as it has unique marker IDs, allowing a fleet to land in its own separate places. This contrasts with Fractal markers, which do not offer multiple unique landing pads, and Harco markers, which rely on separate parent and child dictionaries and thus do not preserve a unified ID across scales. 

The remainder of this paper is organized as follows. Section \ref{sec::relworks} reviews the related works on fiducial markers. Section \ref{sec:proposal} details the proposed generation and detection algorithms for Recursive ArUco markers. Section \ref{sec::experiments} presents the experiments and results to validate our approach. Finally, Section \ref{sec::conclusions} draws the main conclusions.

\section{Related Works}
\label{sec::relworks}
Fiducial markers have been extensively studied for their robustness and computational efficiency in various applications, ranging from augmented reality \cite{azuma1997survey} to camera positioning in complex indoor environments \cite{sparse_indoor}. In recent years, vision-based autonomous landing for Unmanned Aerial Vehicles (UAVs) has emerged as a prominent application of fiducial markers, leading to extensive research into novel algorithms and techniques to ensure safe and precise tracking \cite{semerikov2025vision}. Standard marker systems, such as ArUco \cite{arucopaper}, AprilTag \cite{wang2016iros}, and ARTag \cite{fiala2005artag}, provide reliable pose estimation and have been widely adopted for UAV landing aid systems \cite{marut_aruco,lebedev2020accurate}. To improve robustness under diverse operational conditions, several approaches have optimized standard marker tracking using techniques such as deep learning for high-speed fixed-wing drones \cite{yuan2022high}, multiple concentric ellipse fitting \cite{jung2015robust}, height-adaptive tracking strategies \cite{liu2020vision}, ground pattern recognition \cite{wubben2019accurate}, and inner-outer nested marker designs \cite{wang2023autonomous,nguyen2017remote}. Recent advances have also focused on improving the computational efficiency of these standard markers. For instance, ArUco Nano \cite{aruco_nano_paper} introduces a lightweight contour-tracing and sub-pixel sampling method that significantly reduces computational overhead, making it well-suited for resource-constrained devices such as UAVs.

However, standard fiducial markers fundamentally struggle when the camera-to-marker distance varies significantly, as they rely on a fixed physical size. To address this limitation, multi-scale or hierarchical marker designs have been proposed. Fractal markers \cite{Fractal} embed smaller markers within a parent marker, enabling detection at shorter ranges, and have been actively applied to autonomous UAV landing on mobile robotic platforms \cite{anikin2023autonomous}. Other works, such as Embedded ArUco \cite{khazetdinov2021embedded}, have extended this idea for high-precision landing. Nevertheless, these fractal-based approaches rely on the marker's center, making them highly susceptible to occlusion. HArCo \cite{Harco} proposes a different hierarchical structure by embedding child markers within the white cells of the parent marker. While this approach improves transition smoothness, it heavily restricts the number of recursion levels and constrains the available dictionary.

Other approaches have explored extending marker density to improve robustness. For example, ChArUco2 \cite{charuco2} enhances standard calibration boards by embedding markers in both black and white squares, thereby doubling the marker density and providing robustness under partial occlusion. While highly effective for calibration, this checkerboard-based design is not optimized for single-marker localization in UAV landing, where the marker must be seamlessly detected across vastly different scales. 

Our proposed Recursive ArUco (RArUco) marker addresses these challenges by embedding identical markers in both black and white bits at arbitrary recursion depths. Crucially, our detection method leverages a border-sampling strategy that does not require the marker's center to be visible. This guarantees unlimited recursion depth and ensures highly robust tracking regardless of the camera's distance, steep viewing angles, or partial occlusions.

\section{Proposed Method}
\label{sec:proposal}
Our proposed approach modifies the generation and detection pipelines of standard square fiducial markers. 

\subsection{Marker Generation}

Our generation algorithm constructs a recursive ArUco marker by repeatedly embedding the image generated at the previous recursion level into every bit of the same marker.
Let $M$ denote a standard marker from a predefined dictionary (e.g., ArUco \cite{arucopaper}, AprilTag \cite{wang2016iros}), represented as a binary matrix where each element (bit) is either black (0) or white (1), and let $D$ denote the maximum desired recursion depth. The same marker $M$, and therefore the same marker ID, is used at every recursion level.

\begin{figure}[H]
\centering
\includegraphics[width=0.9\linewidth]{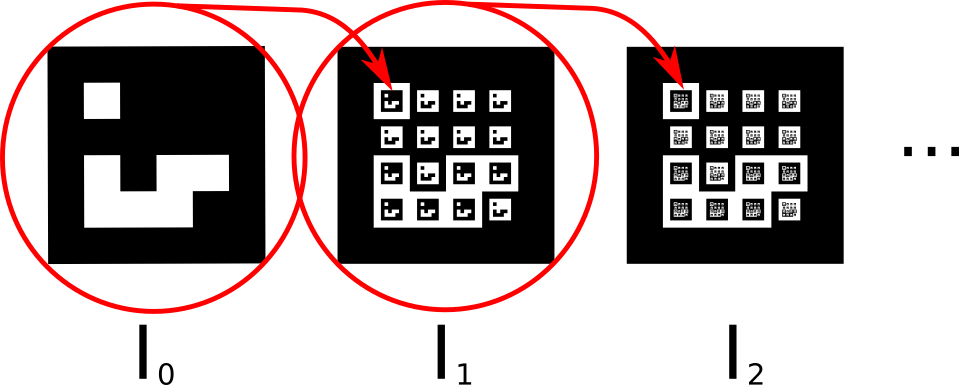}
\caption{Generation process of RArUco markers.  Markers are inserted into the bits of the original marker, adding a surrounding border. Colors are inverted when inserted in black pixels.}
\label{fig:markerGeneration}
\end{figure}

The image $I_0$ corresponds to the standard, non-recursive representation of $M$. For each recursion depth $d \in {1,\ldots,D}$, the image $I_d$ is constructed according to the bit pattern of $M$ by embedding the image from the previous recursion level, $I_{d-1}$, into every bit cell (see Figure \ref{fig:markerGeneration}).

To preserve the marker's visual structure at each scale and improve its detectability, the embedded image does not occupy the entire area allocated to its parent bit. Instead, a border of width $b$, having the same color as the parent bit, is retained around the embedded image. Consequently, the spatial resolution of $I_d$ increases with the recursion depth to accommodate both the copies of $I_{d-1}$ and their surrounding borders.

Formally, for each recursion depth $d \in {1,\ldots,D}$ and every bit location $(i,j)$, the corresponding region of $I_d$ is generated as follows:
\begin{enumerate}
\item If the bit $M_{i,j}$ is white, a copy of $I_{d-1}$ is placed at the center of the allocated region and surrounded by a white border of width $b$.
\item If the bit $M_{i,j}$ is black, an inverted copy $\overline{I_{d-1}}$, obtained by exchanging black and white pixels, is placed at the center of the allocated region and surrounded by a black border of width $b$.
\end{enumerate}

Although the generation procedure could be generalized to embed a different marker at each recursion level, our implementation uses the same marker ID throughout the entire recursive structure. This design is particularly suitable for multi-UAV landing-pad applications. Since each UAV is assigned a landing pad with a specific marker ID, using the same ID at all scales allows the UAV to identify its target unambiguously, regardless of the recursion level at which the marker is detected. Consequently, the total number of distinct landing pads is equal to the number of unique markers available in the selected dictionary.

We additionally recommend using dictionaries with relatively small bit grids, such as $4 \times 4$ or $5 \times 5$. For a marker of fixed physical size, a smaller grid results in larger individual bits, improving their visibility and increasing the potential detection distance. The border width $b$ also plays an important role in detection robustness. Increasing $b$ increases the visible area associated with each parent bit, which can improve detection at steep viewing angles and under strong perspective distortions. However, this comes at the cost of reducing the area available for the recursively embedded marker.

\subsection{Marker Detection}
\begin{figure}[t!]
\centering
\includegraphics[width=1\linewidth]{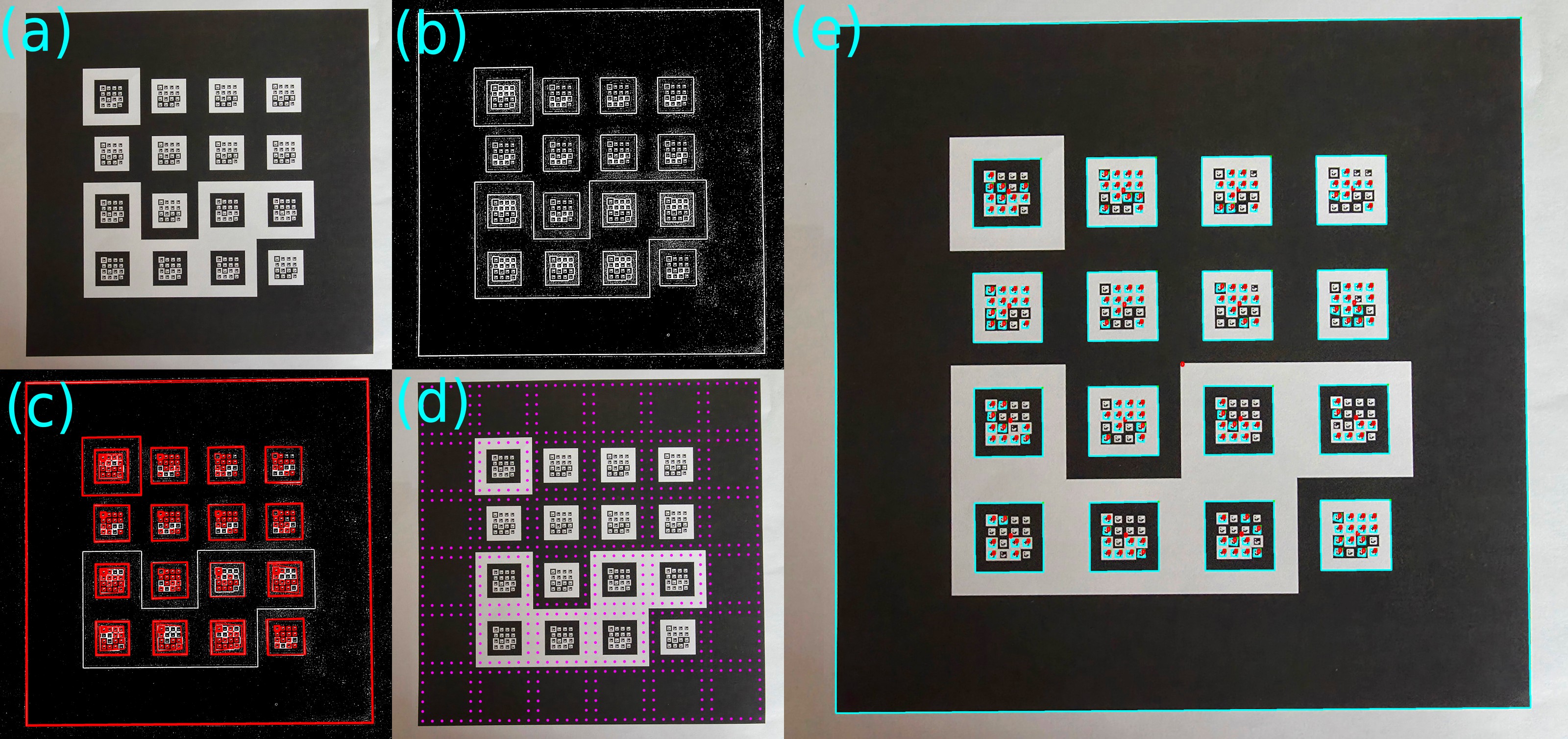}
\caption{Steps of RArUco detection. (a) Input image. (b) Adaptive threshold. (c)  Rectangles obtained after contour extraction. (d) Bit color estimation by sampling a grid of inner rectangles for each marker bit. Only the samples for the most external rectangle are shown. (e) Detected $181$ out of the $273$ total markers in the image. Only markers of the deepest levels are not detected in this image. }
\label{fig:detection}
\end{figure}

The detection of standard ArUco markers typically involves adaptive segmentation, contour extraction, polygonal approximation to find square candidates, and finally, perspective removal to extract the bit matrix. RArUco marker detection follows the same general pipeline, with specific adaptations to account for its recursive structure (see Figure \ref{fig:detection}). To minimize computational bottlenecks and enable high-speed processing on resource-constrained devices, our approach integrates highly optimized techniques inspired by ArUco Nano \cite{aruco_nano_paper}. Specifically, we utilize lightweight high-pass filtering for segmentation (Figure \ref{fig:detection}(b)) and a Visited-Aware contour tracing algorithm that efficiently rejects noise artifacts without relying on expensive standard contour-finding functions (Figure \ref{fig:detection}(c)).

To determine the color of each bit, standard implementations compute the median or mean intensity of all pixels within the bit's cell, or sample the central pixel. These standard sampling strategies fail for our Recursive ArUco markers because the center of every bit is populated by the inner marker $I_{d-1}$, which contains both black and white pixels. 

To overcome this, we introduce a modified bit extraction strategy. Once a square candidate is localized and rectified, the color of each bit is determined by sampling only the pixels residing in the border region of width $b$, explicitly ignoring the central region (Figure \ref{fig:detection}(d)). Mathematically, for a given bit cell $C_{i,j}$ defined over the domain $[0, S] \times [0, S]$ (where $S$ is the bit size in pixels), we define a sampling mask $W$ that excludes the central area $[b, S-b] \times [b, S-b]$. The bit intensity is then computed over the region $C_{i,j} \cap W$. Moreover, instead of employing costly full-image warping via homography, our detection method uses direct sub-pixel code sampling with stochastic corner refinement to robustly extract bit values (as proposed in \cite{aruco_nano_paper}).

Furthermore, because markers embedded within black bits are inverted during generation, the detection algorithm is adapted to recognize both standard (black-and-white) and inverted (white-and-black) markers. During the contour extraction phase, the system identifies both dark square candidates on light backgrounds and light square candidates on dark backgrounds. If an inverted candidate is processed, its bit intensities are logically reversed before decoding.

This border-sampling technique is the key to our method's robustness. Because the algorithm only inspects the perimeter of each bit to ascertain its value, it is entirely agnostic to the contents of the bit's center. Consequently, any inner marker---or indeed any arbitrary occlusion at the center of the marker---will not disrupt detection of the parent marker. This provides a significant advantage over other approaches, which mandate center visibility.

\section{Experiments and Results}
\label{sec::experiments}
To validate the efficacy of the proposed Recursive ArUco markers, we conducted a series of experiments focusing on five critical aspects: operational distance range and viewing angles, robustness to occlusion, robustness to partial visibility, robustness to motion blur, and real-world flight test applicability.  Note that while Embedded ArUco and Harco markers also employ nested hierarchical structures, they were excluded from our empirical baseline evaluation. Neither the source code nor pre-compiled executables for these systems are publicly available for testing. Because their internal generation and detection pipelines rely on specific algorithmic optimizations not fully detailed in the literature, an independent re-implementation could inadvertently misrepresent their true performance. Furthermore, Fractal markers already serve as a strong representative baseline for center-dependent multi-scale approaches. This renders any direct, quantitative comparison with unavailable methods scientifically unfair and potentially biased. To ensure strict reproducibility and a rigorous baseline, we restricted our comparative analysis to widely adopted, publicly verifiable frameworks: the standard ArUco detector implemented in OpenCV 4.12, and the original Fractal marker implementation (\url{https://sourceforge.net/projects/aruco/}). The proposed generation and detection algorithms were integrated into our new ArUco2 library, whose source code is publicly available at \url{https://github.com/rmsalinas/aruco2}. All computational performance tests were carried out on a system equipped with a 13th Gen Intel(R) Core(TM) i7-13700H processor. Finally, we must indicate that in all our tests, we employed markers using $b=2$, which, according to our experience, obtained a good trade-off between angle and distance detection.

\subsection{Distance and Viewing Angle Evaluation}
The primary motivation for recursive markers is to enable continuous detection over varying camera-to-target distances. While UAV landing scenarios primarily involve relatively perpendicular approaches, we additionally evaluate the methods under steep viewing angles. Although such extreme angles may not be strictly necessary for standard landing pads, this evaluation demonstrates the overall robustness of our proposal and its potential for more general-purpose applications.

\begin{figure}[htbp]
\centering
\begin{minipage}{0.32\textwidth}
  \centering
  \includegraphics[width=\linewidth]{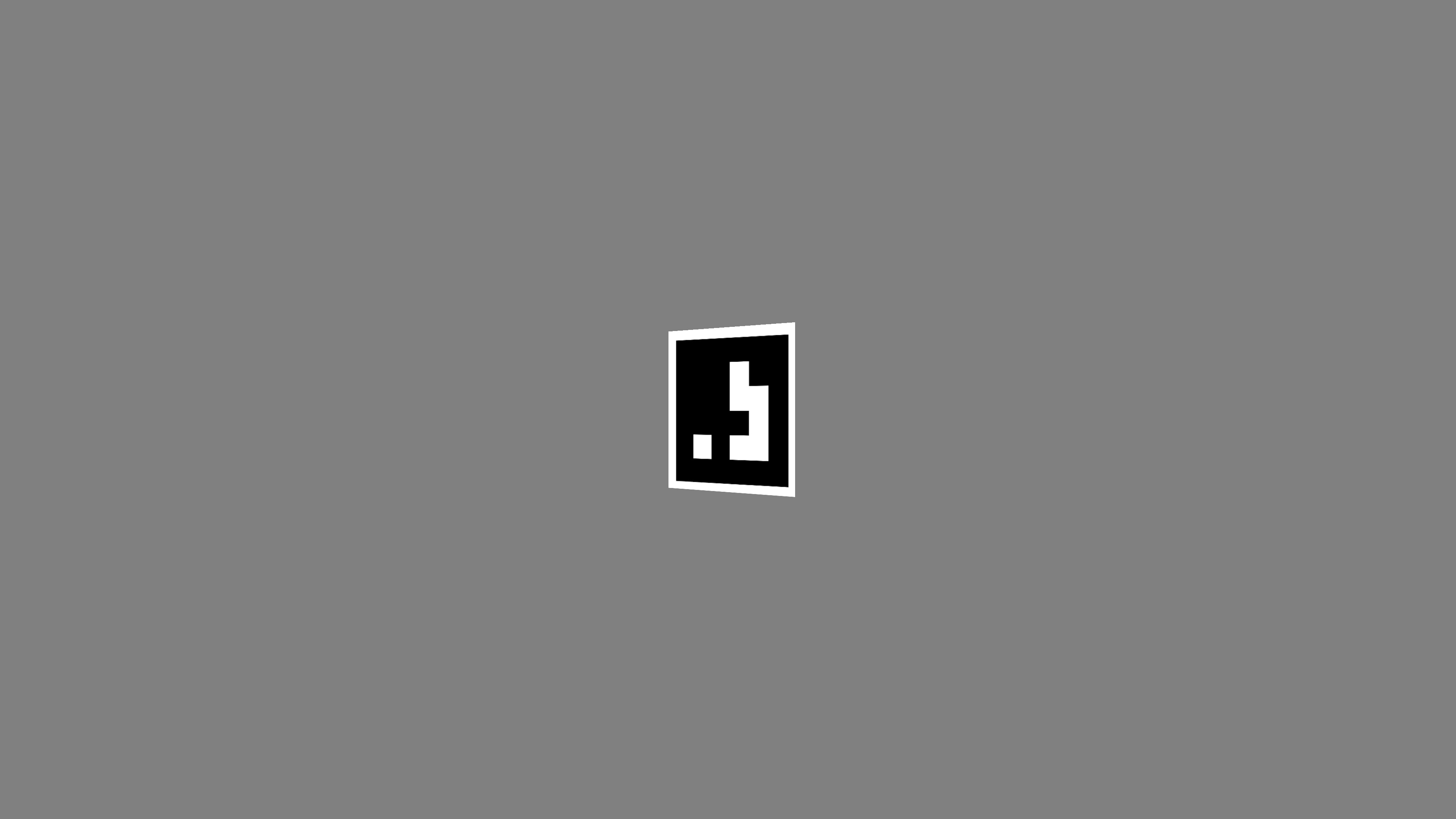}
  \caption*{Marker}
\end{minipage}%
\hfill
\begin{minipage}{0.32\textwidth}
  \centering
  \includegraphics[width=\linewidth]{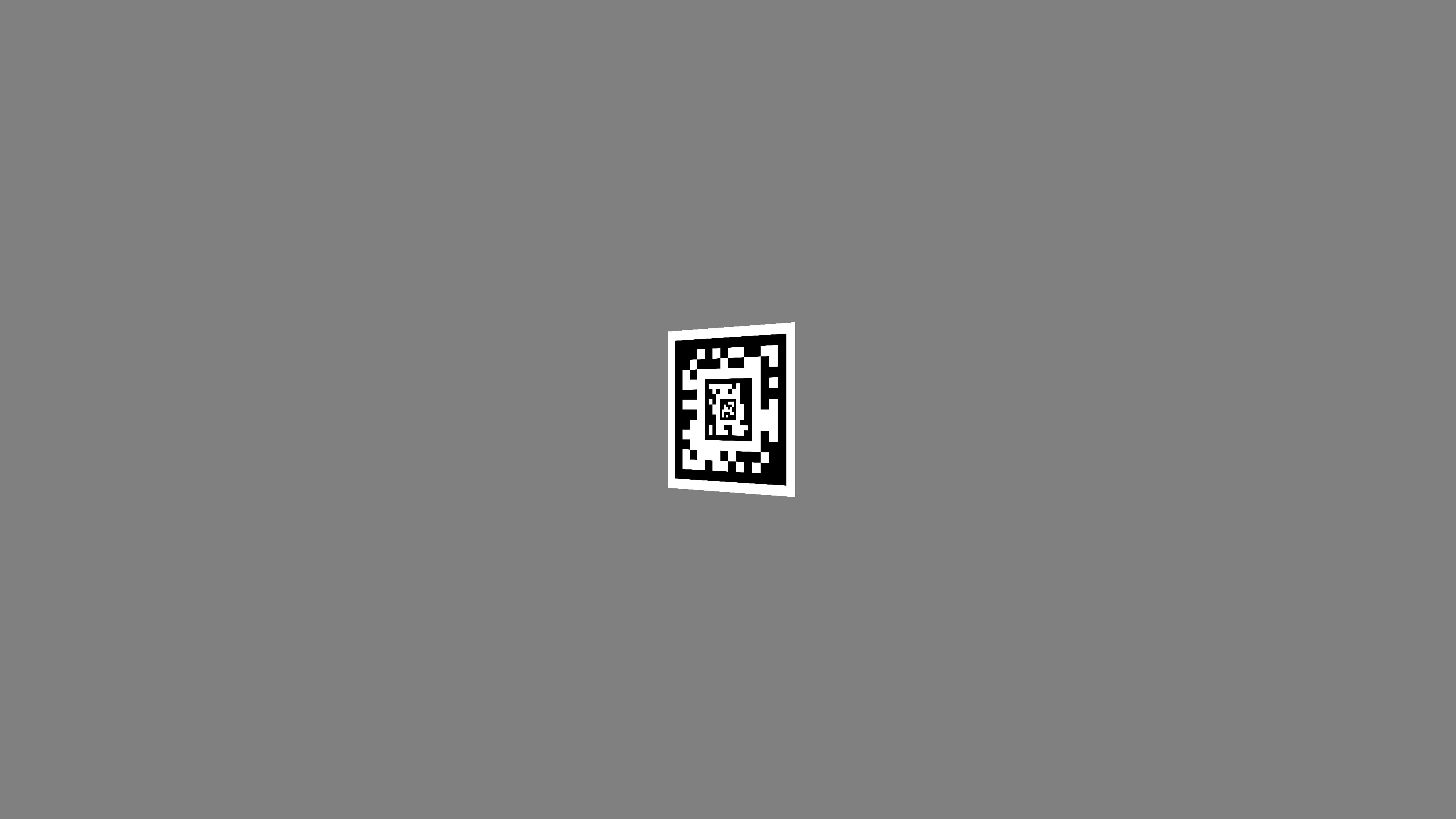}
  \caption*{Fractal}
\end{minipage}%
\hfill
\begin{minipage}{0.32\textwidth}
  \centering
  \includegraphics[width=\linewidth]{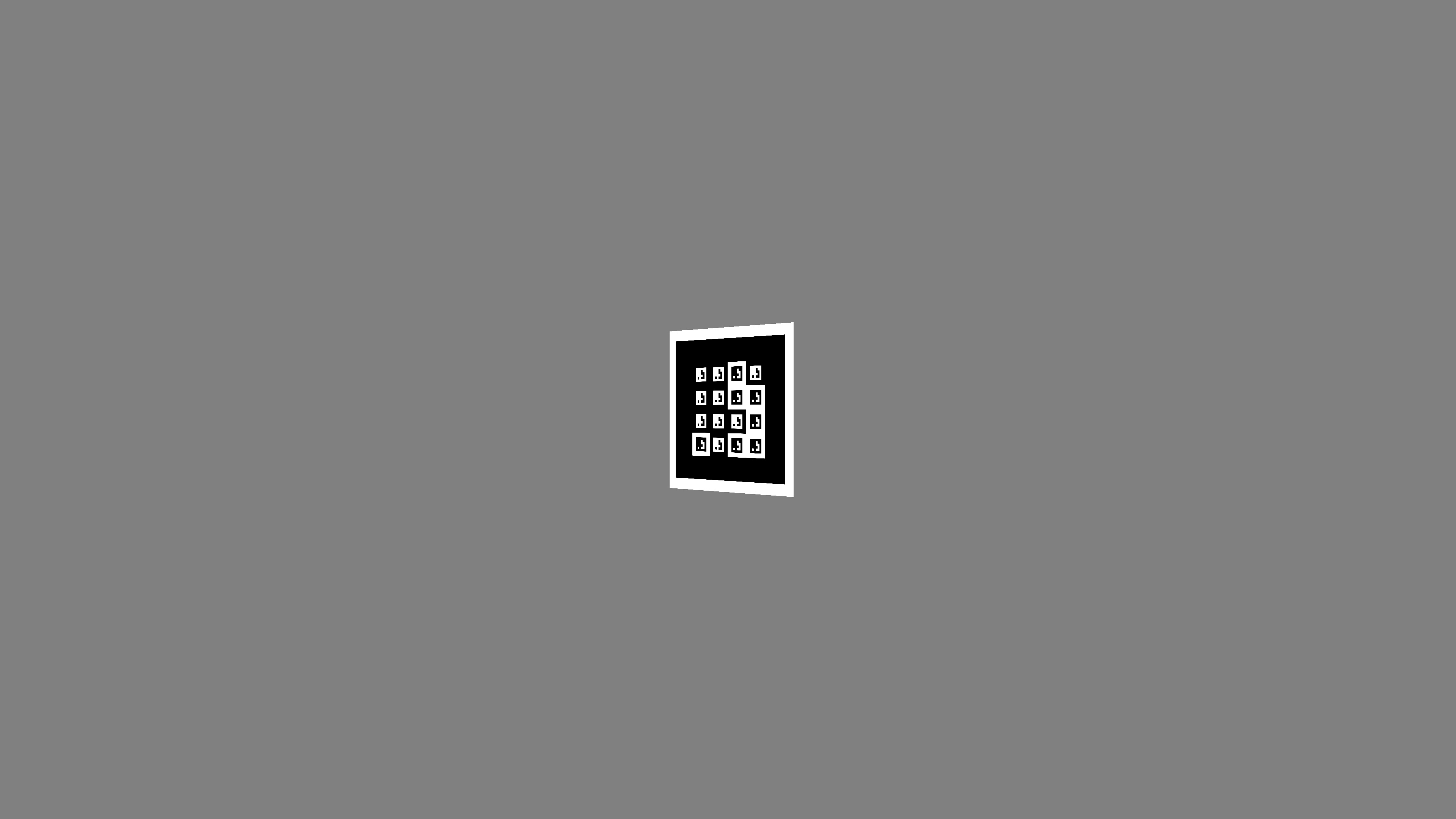}
  \caption*{RArUco}
\end{minipage}
\caption{Examples from the simulated dataset used for evaluation. The images depict the standard ArUco, Fractal, and RArUco markers viewed from a distance of 5.74 meters at a $40^\circ$ angle.}
\label{fig:simulated_examples}
\end{figure}

\textbf{Experimental Setup:} To benchmark the detection capabilities of different marker methods under varying viewpoints, we developed a simulation framework. This framework generates simulated 4K (3840$\times$2160) images of a marker placed in a 3D environment, captured from a virtual camera with a standard focal length and light lens distortion. We evaluated three types of fiducial markers (each modeled with a physical size of 1.0 $\times$ 1.0 meters): standard ArUco markers (evaluated using the DICT\_APRILTAG\_16h5 dictionary), Fractal markers (specifically FRACTAL\_3L\_6), and our proposed Recursive ArUco (RArUco) markers (generated with 2 recursion levels using marker 0 of the DICT\_APRILTAG\_16h5 dictionary). The camera position was systematically varied across a grid of distances (ranging from 0.5 to 100 meters) and viewing angles (ranging from $0^\circ$ to $80^\circ$ with respect to the marker's normal). Figure~\ref{fig:simulated_examples} shows a representative example of the three marker types captured from a distance of 5.74 meters at a $40^\circ$ viewing angle.

\begin{figure}[htbp]
\centering
\begin{minipage}{0.32\textwidth}
  \centering
  \captionsetup{justification=centering}
  \includegraphics[width=\linewidth]{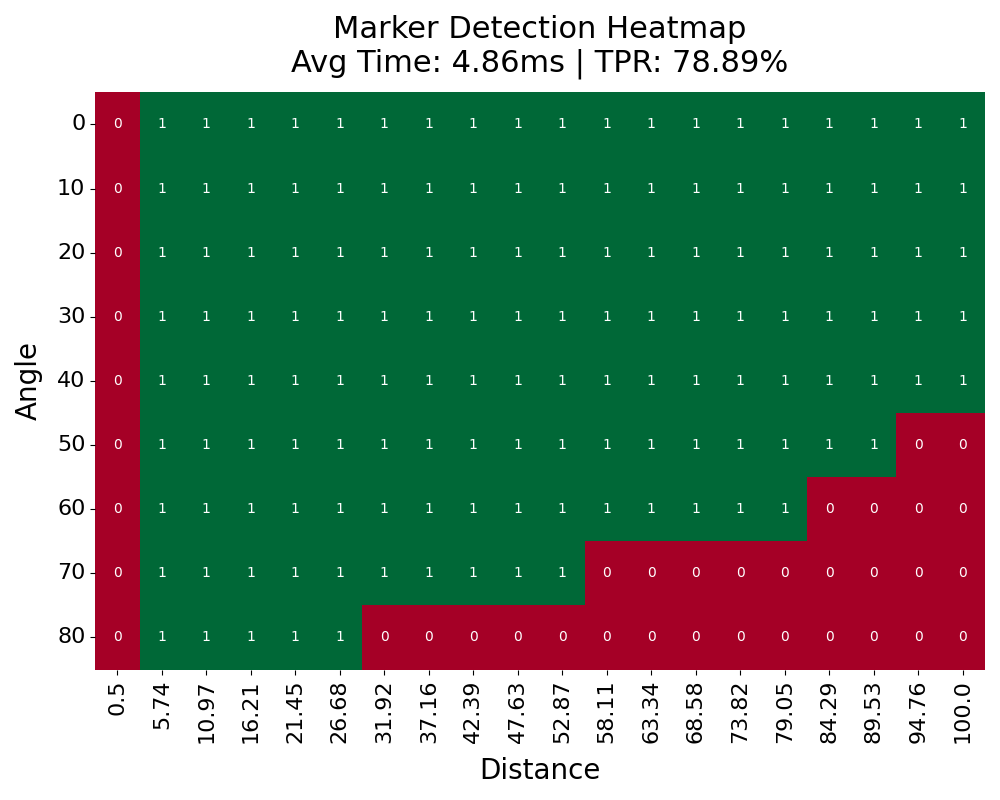}
  \caption*{Marker}
\end{minipage}%
\hfill
\begin{minipage}{0.32\textwidth}
  \centering
  \captionsetup{justification=centering}
  \includegraphics[width=\linewidth]{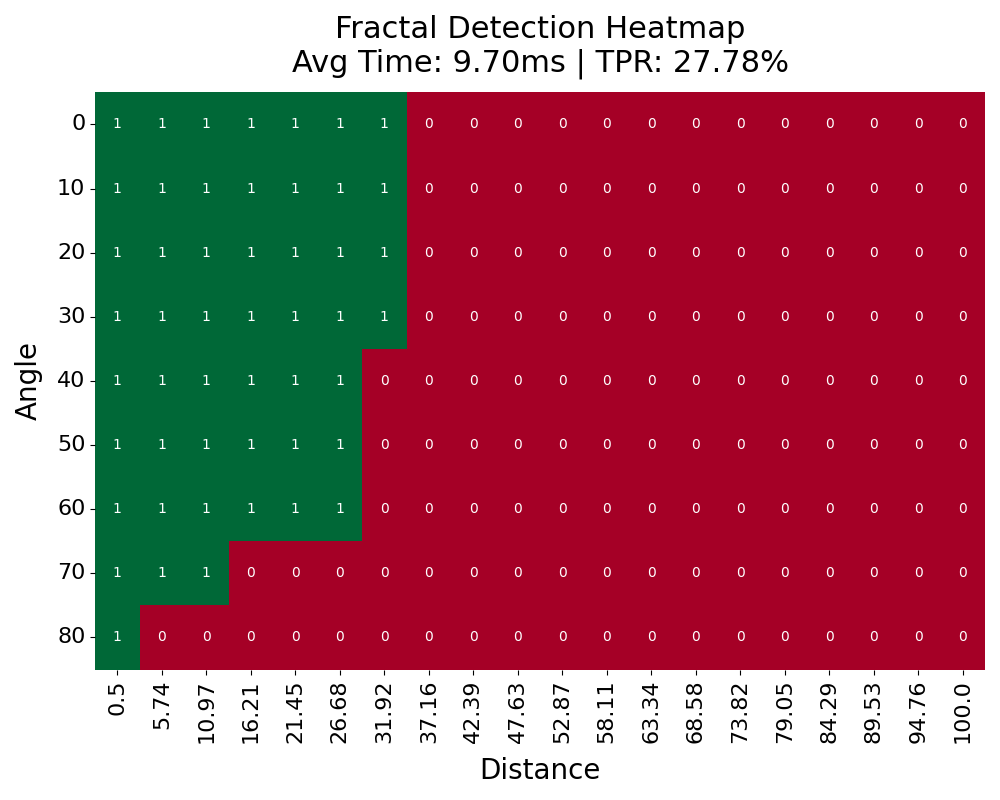}
  \caption*{Fractal}
\end{minipage}%
\hfill
\begin{minipage}{0.32\textwidth}
  \centering
  \captionsetup{justification=centering}
  \includegraphics[width=\linewidth]{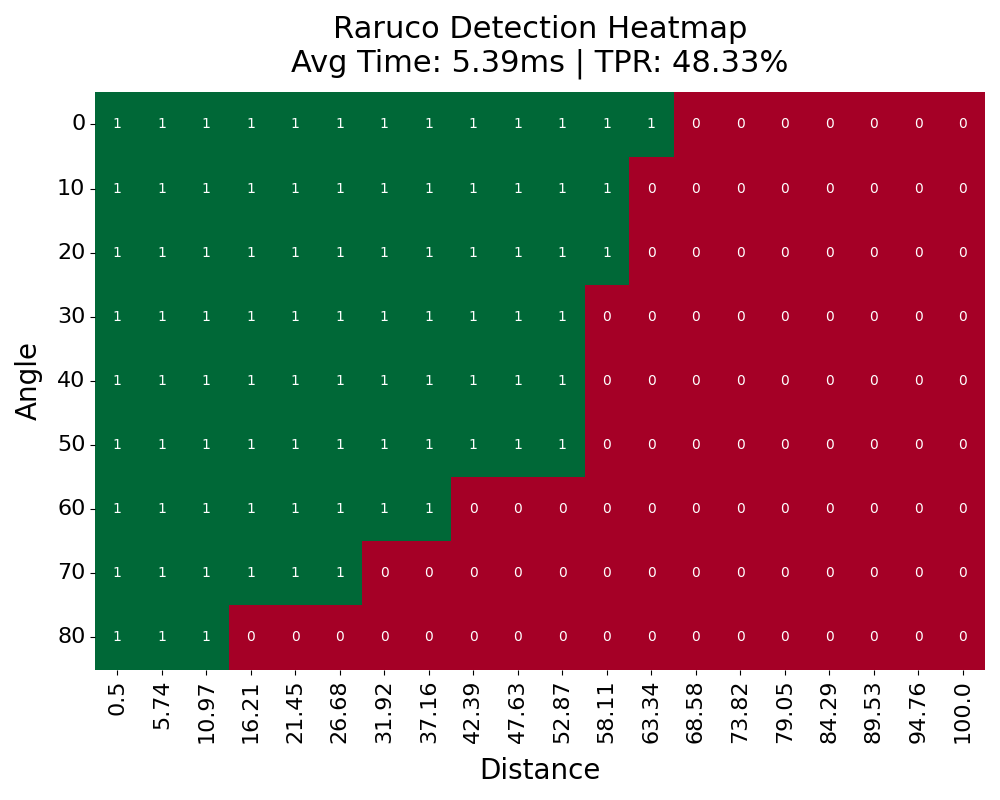}
  \caption*{RArUco}
\end{minipage}
\caption{True Positive Rate (TPR) heatmaps for standard ArUco, Fractal, and RArUco markers across varying camera distances (x-axis) and viewing angles (y-axis).}
\label{fig:heatmaps}
\end{figure}

\textbf{Results:} The True Positive Rate (TPR) and average computing times were calculated for each method across the simulated viewpoint grid.  Figure~\ref{fig:heatmaps} shows the results as heatmaps, where each cell represents a specific camera distance and viewing angle; a value of 1 (green) indicates successful marker detection, while a 0 (red) denotes failure.  The standard ArUco marker exhibits a high overall detection rate (78.89\%) and an average computation time of 11.69 ms. Still, it inherently fails at close range when the marker borders extend beyond the camera's field of view. The Fractal marker, while designed for robustness, achieved an overall TPR of only 27.78\% (average computing time of 13.10 ms) and showed particular vulnerability to steep viewing angles. Conversely, our proposed RArUco markers achieved a competitive overall TPR of 48.33\% (average computing time of 5.39 ms). Crucially, the RArUco marker seamlessly maintained detection at extremely close distances by transitioning to its embedded inner markers, while remaining highly robust to extreme viewing angles up to $80^\circ$, significantly outperforming the Fractal marker in challenging perspective conditions. In particular, our method demonstrated a substantially larger operational distance range than the Fractal marker, maintaining detection up to approximately 63.3 meters compared to the Fractal marker's limit of roughly 31.9 meters (at a $0^\circ$ viewing angle). Furthermore, operating at approximately 185 FPS, RArUco is significantly faster than both the standard ArUco marker ($\sim$85 FPS) and the Fractal marker ($\sim$76 FPS), making it highly suitable for real-time UAV control loops.

\subsection{Robustness to Occlusion}
A significant advantage of our approach over existing recursive designs (such as Fractal markers) is its independence of the marker's center. To empirically validate this, we conducted a rigorous occlusion experiment comparing standard ArUco markers, Fractal markers, and our proposed Recursive ArUco (RArUco) markers.

\begin{figure}[htbp]
\centering
\begin{minipage}{0.32\textwidth}
  \centering
  \captionsetup{justification=centering}
  \includegraphics[width=\linewidth]{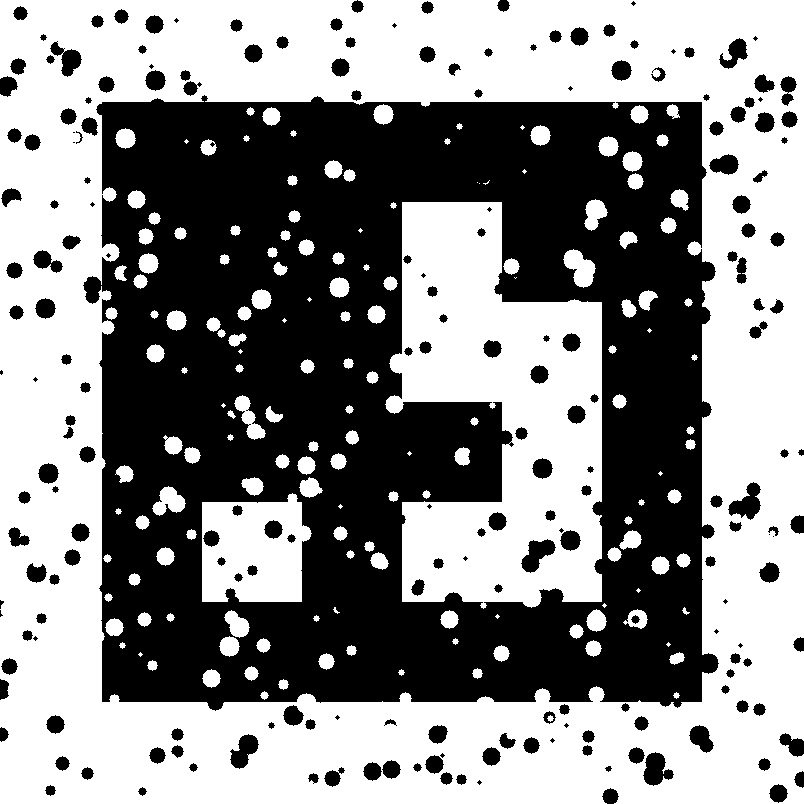}
  \caption*{Marker (15\%)}
\end{minipage}%
\hfill
\begin{minipage}{0.32\textwidth}
  \centering
  \captionsetup{justification=centering}
  \includegraphics[width=\linewidth]{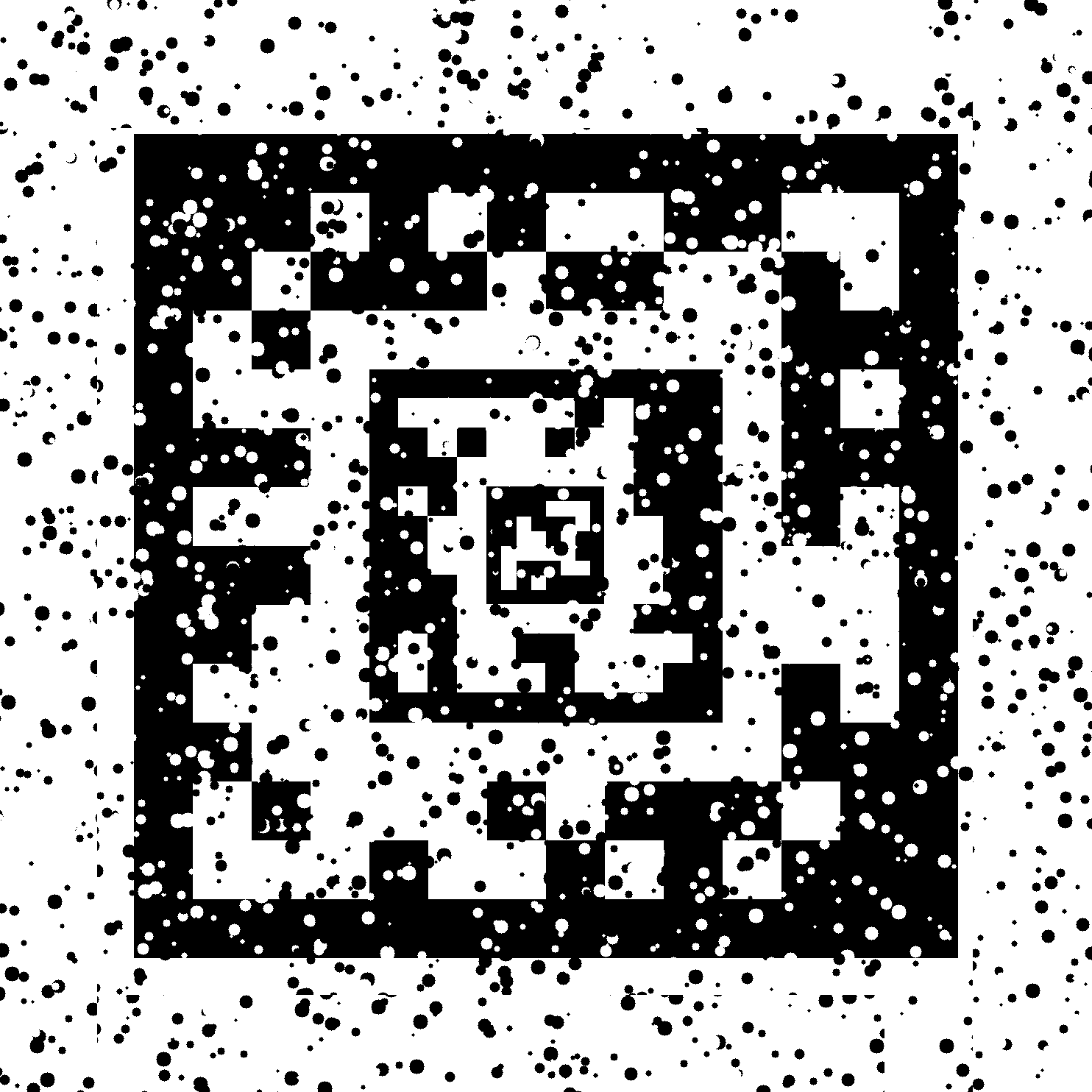}
  \caption*{Fractal (15\%)}
\end{minipage}%
\hfill
\begin{minipage}{0.32\textwidth}
  \centering
  \captionsetup{justification=centering}
  \includegraphics[width=\linewidth]{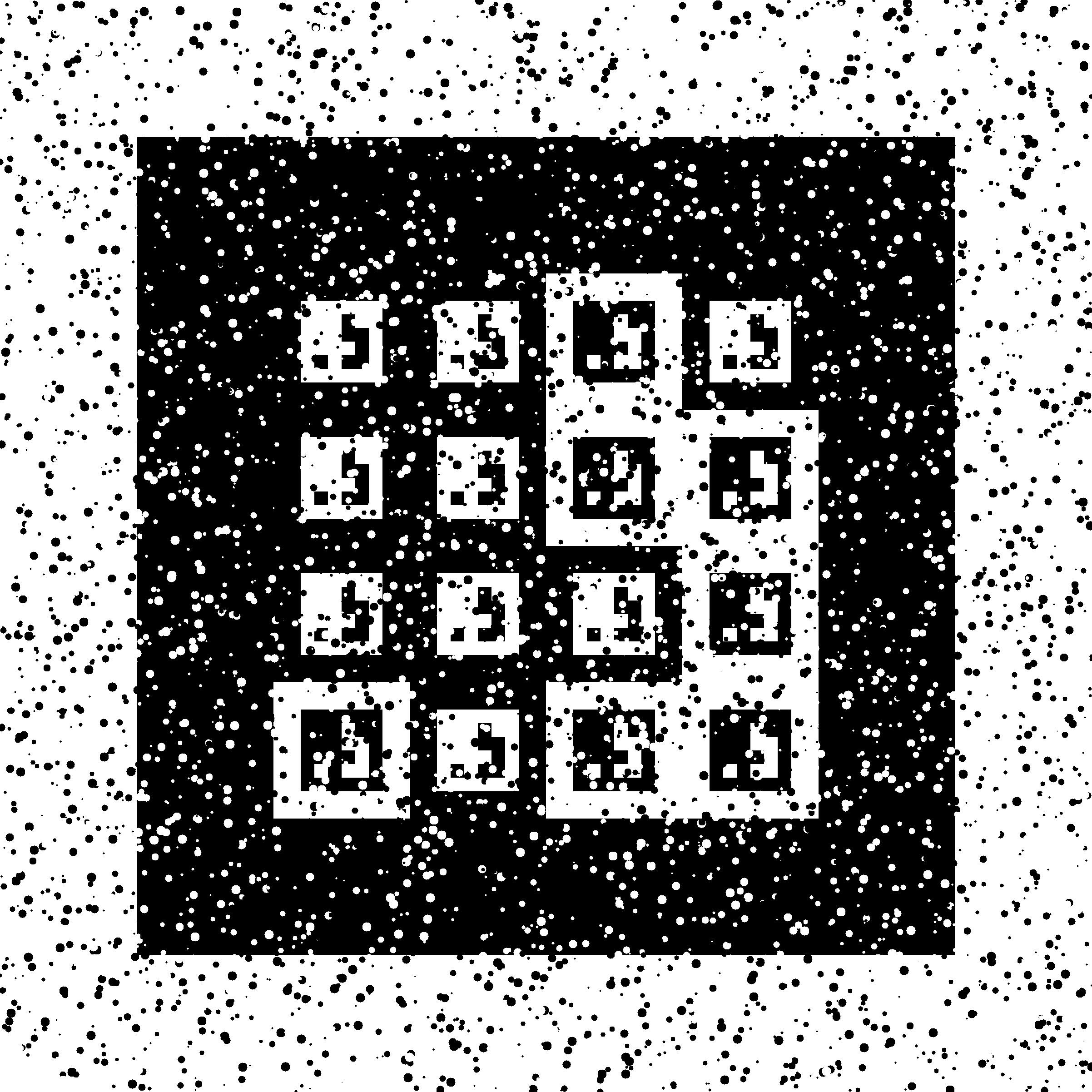}
  \caption*{RArUco (15\%)}
\end{minipage}
\caption{Examples of the simulated occlusion noise applied to the three marker types at a 15\% occlusion level.}
\label{fig:occlusion_samples}
\end{figure}

\textbf{Experimental Setup:} Each marker type was generated and padded with a white boundary to simulate a real-world scenario. We then systematically applied random occlusion noise to the images. The noise consisted of randomly placed black-and-white circles (with radii ranging from 2 to 10 pixels). We evaluated occlusion levels ranging from 5\% to 80\% of the total image area. For each occlusion level and marker type, we performed 100 independent trials, ensuring that the noise generator advanced its random state uniformly across all trials. Representative samples of the markers with 15\% occlusion noise are shown in Figure~\ref{fig:occlusion_samples}.

\begin{figure}[htbp]
\centering
\includegraphics[width=0.8\linewidth]{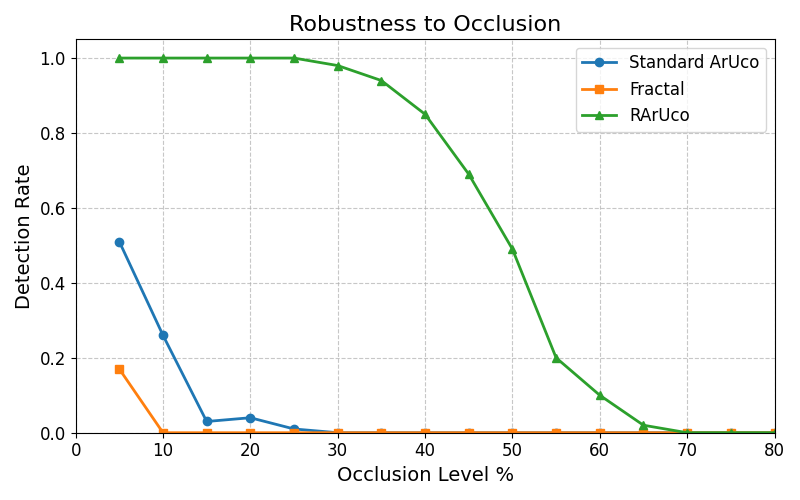}
\caption{Detection rates for standard ArUco, Fractal, and RArUco markers as a function of the random occlusion noise level. Our proposed RArUco maintains a 100\% detection rate up to 30\% occlusion.}
\label{fig:robustness_occlusion}
\end{figure}

\textbf{Results:} The detection rates for each method across the tested occlusion levels are shown in Figure~\ref{fig:robustness_occlusion}. Standard ArUco markers are highly sensitive to occlusions that break their outer square contour, leading to a rapid decline in detection performance even at 5\%-10\% occlusion. The Fractal marker also exhibited severe performance degradation under these conditions, largely because it mandates the visibility of its central sub-markers for successful identification. Conversely, our proposed RArUco markers demonstrated exceptional robustness. By sampling only the bit borders and leveraging stochastic bit extraction, the RArUco marker maintained a 100\% detection rate up to 30\% random occlusion, degrading gracefully only as the noise level exceeded 40\%. Note that in these extreme degradation scenarios, 'detection' refers to successfully decoding any valid RArUco ID within the hierarchy. When the parent marker is compromised, the detection seamlessly falls back to a fully visible inner sub-marker.

\subsection{Robustness to Partial Visibility}
In practical UAV landing scenarios, the drone may not always be perfectly centered over the landing pad, so the camera view captures only a portion of the marker. To evaluate performance under such conditions, we conducted a partial-visibility experiment. 

\begin{figure}[htbp]
\centering
\begin{minipage}{0.22\textwidth}
  \centering
  \includegraphics[width=\linewidth]{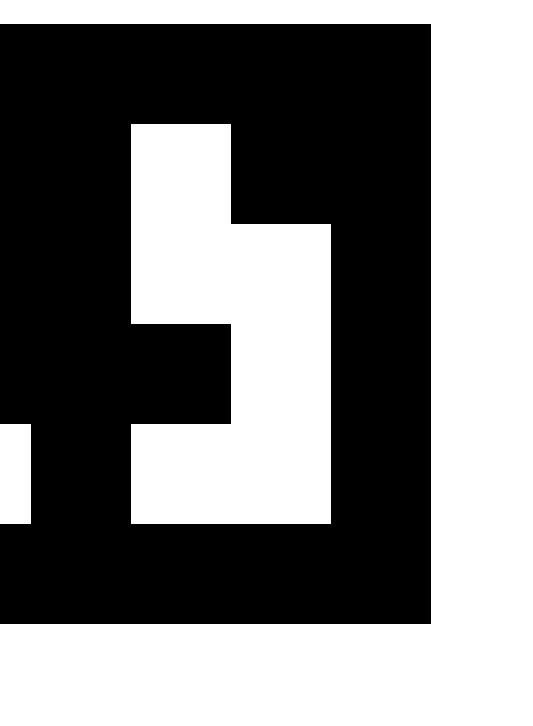}
  \caption*{Marker (40\% Shift)}
\end{minipage}%
\hfill
\begin{minipage}{0.22\textwidth}
  \centering
  \includegraphics[width=\linewidth]{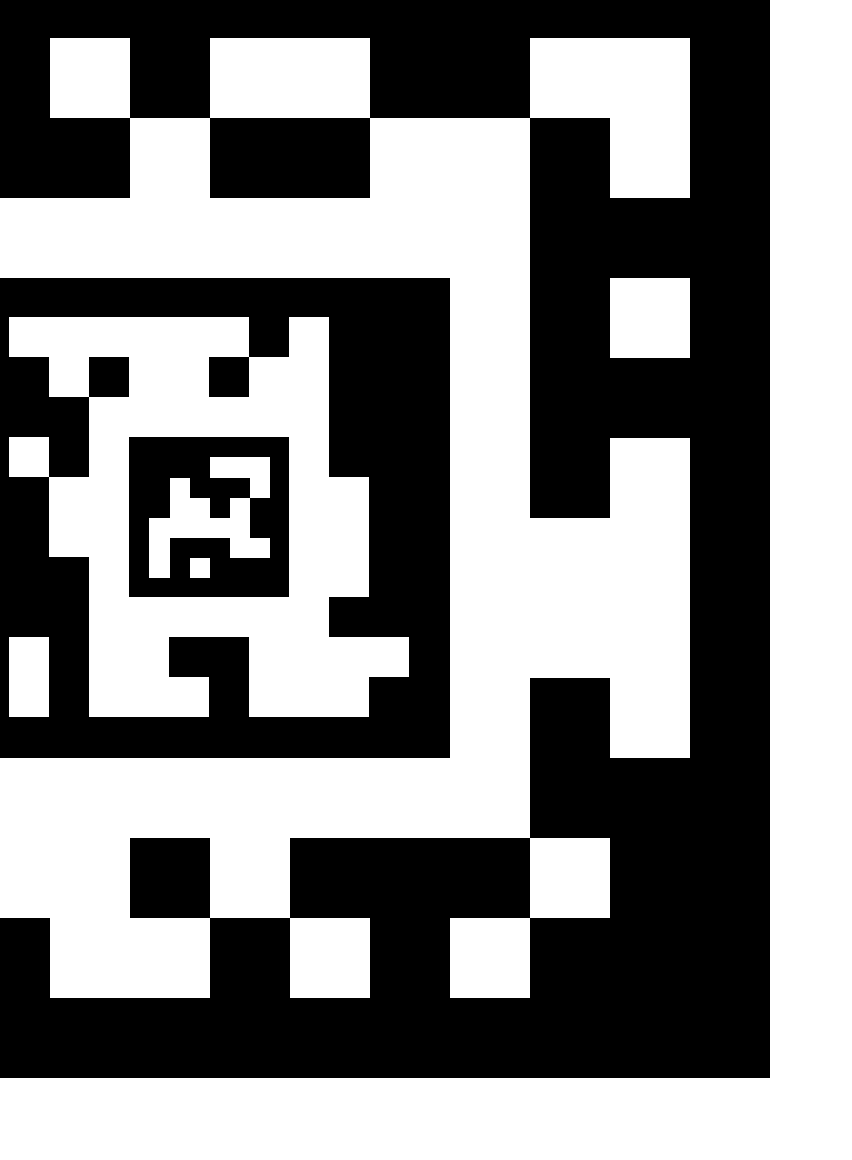}
  \caption*{Fractal (40\% Shift)}
\end{minipage}%
\hfill
\begin{minipage}{0.22\textwidth}
  \centering
  \includegraphics[width=\linewidth]{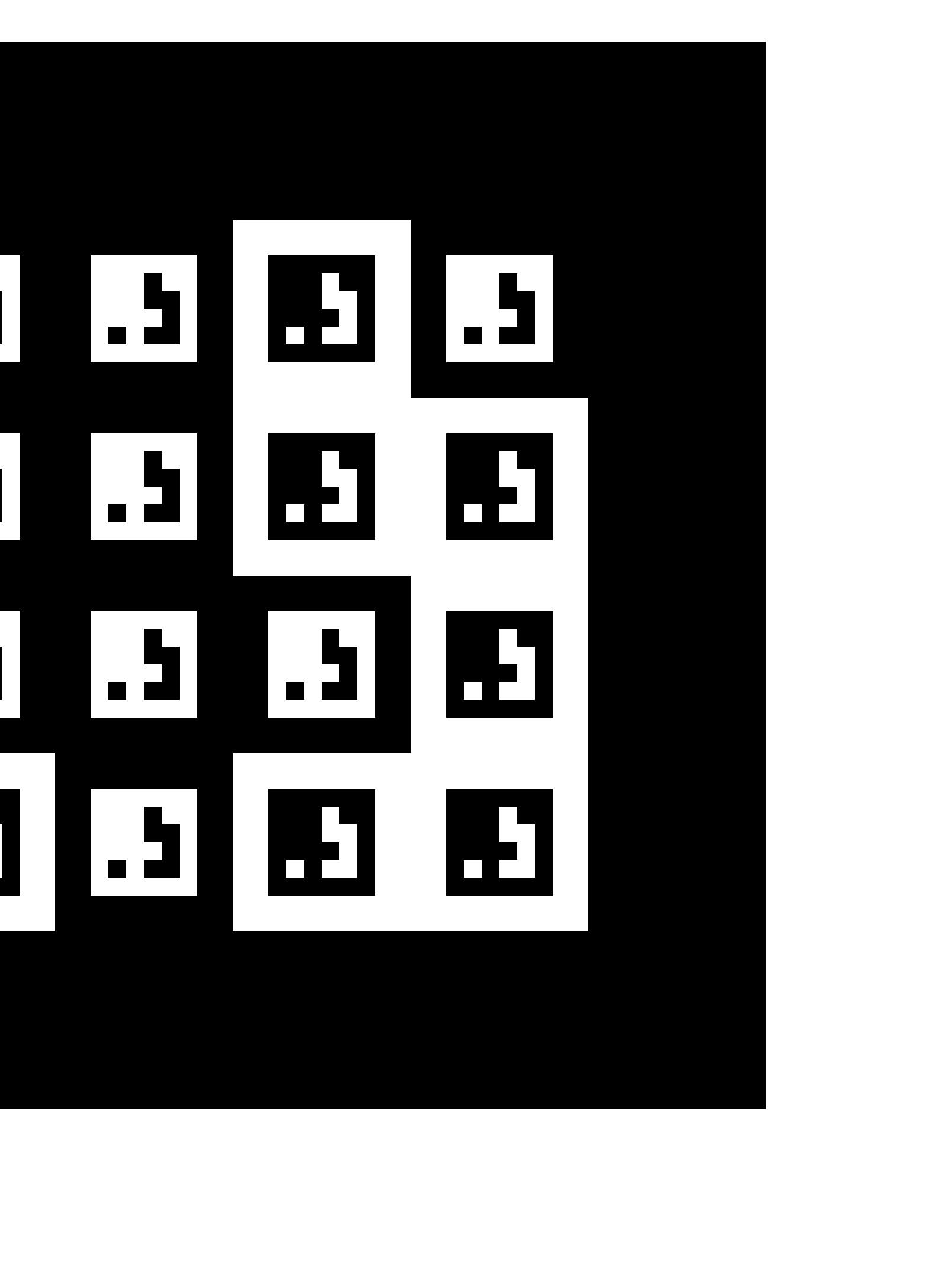}
  \caption*{RArUco (40\% Shift)}
\end{minipage}
\caption{Examples of markers under partial visibility, shown with a 40\% shift (60\% visible ratio).}
\label{fig:shift_samples}
\end{figure}

\textbf{Experimental Setup:} We systematically shifted the markers out of the camera's field of view in completely random directions (angles). The shift magnitude was parameterized as the percentage of the marker's area shifted outside the frame, ranging from 0\% to 100\% (i.e., from 100\% to 0\% visible). For each shift level, we averaged the detection rates across 100 independent random trials. In practical applications, approximately 20\% of the marker's total area corresponds to the required outer white border (quiet zone), meaning the marker itself is unaffected until the shift exceeds this threshold. Figure~\ref{fig:shift_samples} displays the marker appearances at a 40\% shift (60\% visible ratio).

\begin{figure}[htbp]
\centering
\includegraphics[width=0.8\linewidth]{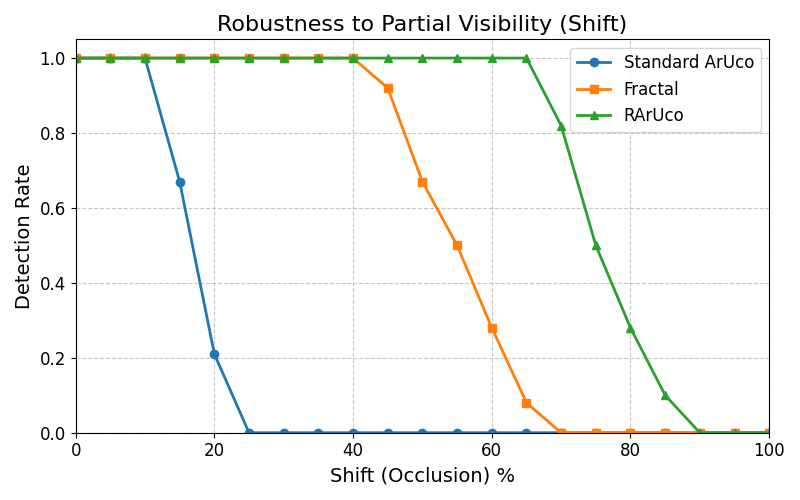}
\caption{Detection rates for standard ArUco, Fractal, and RArUco markers under varying levels of partial visibility. The RArUco marker remains fully detectable even when 60\% of its area is cropped.}
\label{fig:robustness_shift}
\end{figure}

\textbf{Results:} Figure~\ref{fig:robustness_shift} illustrates the robustness of each marker type to partial visibility. Standard ArUco markers are strictly constrained by their outer boundaries; consequently, their detection rate begins to decline immediately after the 20\% white border is shifted out, plummeting to 0\% at a 25\% shift. Fractal markers offer slightly better resilience, sustaining detection up to a 40\% shift, provided that one of their complete inner sub-markers remains fully visible. RArUco markers, however, significantly outperform both alternatives, maintaining a 100\% detection rate even at a 65\% shift. This effectively means that even if a drone drifts such that nearly two-thirds of the landing pad is cut out of the camera frame, RArUco still enables successful tracking and landing.

\subsection{Robustness to Motion Blur}
UAV landings are frequently characterized by sudden maneuvers, vibrations, and high-speed descents, which introduce severe motion blur into the camera feed. To assess the reliability of our markers under these dynamic conditions, we simulated linear motion blur.

\begin{figure}[htbp]
\centering
\begin{minipage}{0.32\textwidth}
  \centering
  \captionsetup{justification=centering}
  \includegraphics[width=\linewidth]{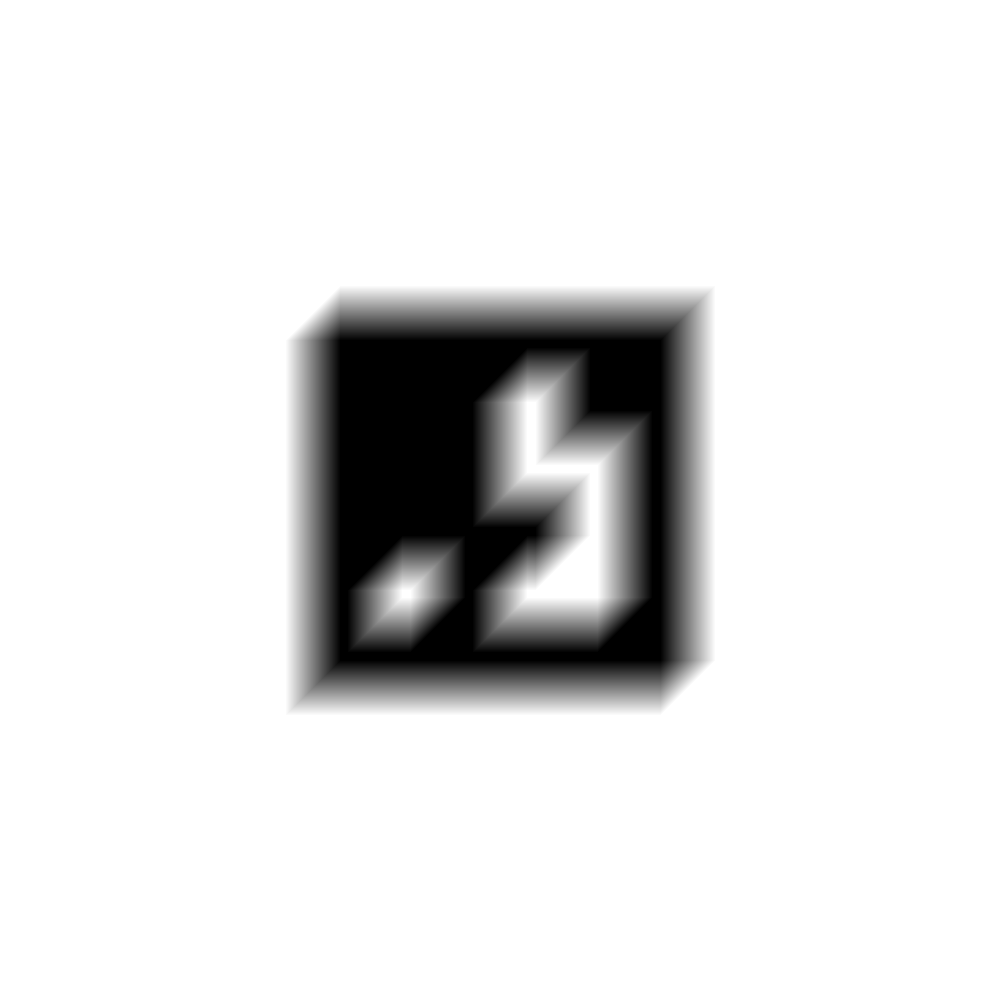}
  \caption*{Marker (15\% blur)}
\end{minipage}%
\hfill
\begin{minipage}{0.32\textwidth}
  \centering
  \captionsetup{justification=centering}
  \includegraphics[width=\linewidth]{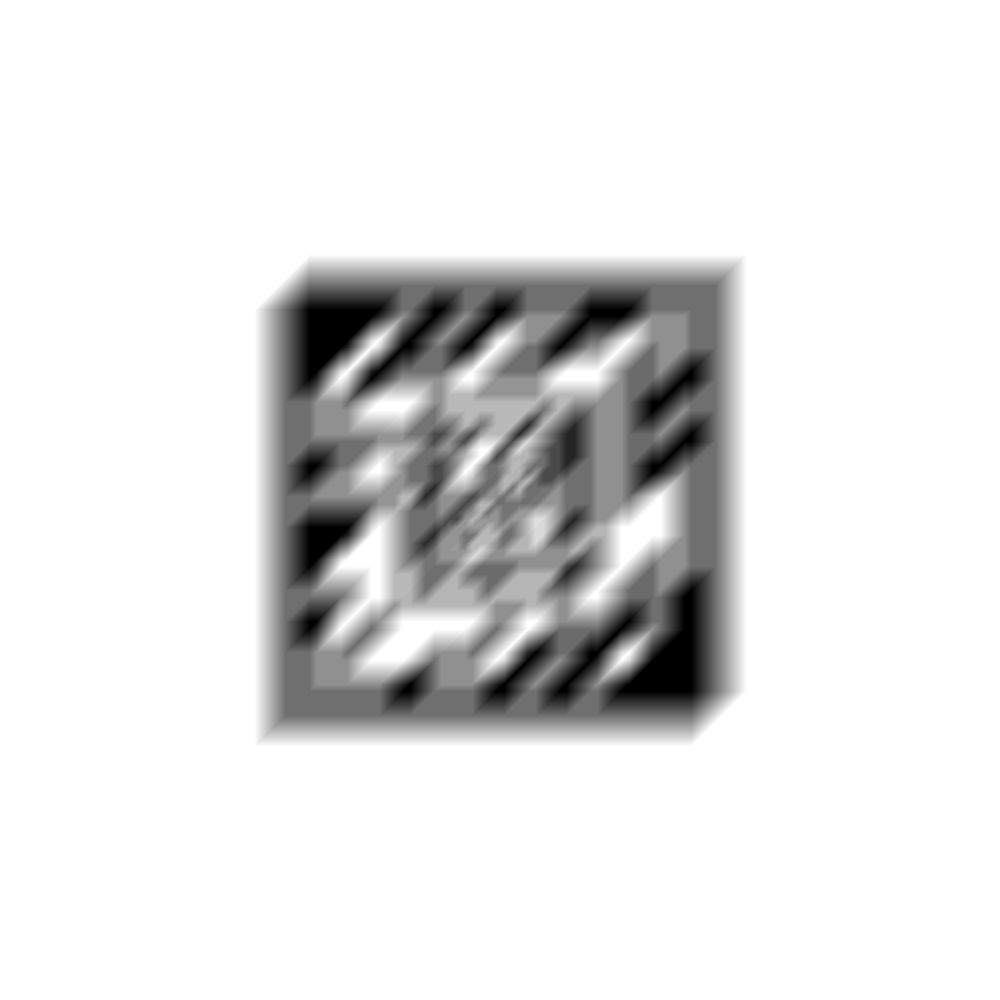}
  \caption*{Fractal (15\% blur)}
\end{minipage}%
\hfill
\begin{minipage}{0.32\textwidth}
  \centering
  \captionsetup{justification=centering}
  \includegraphics[width=\linewidth]{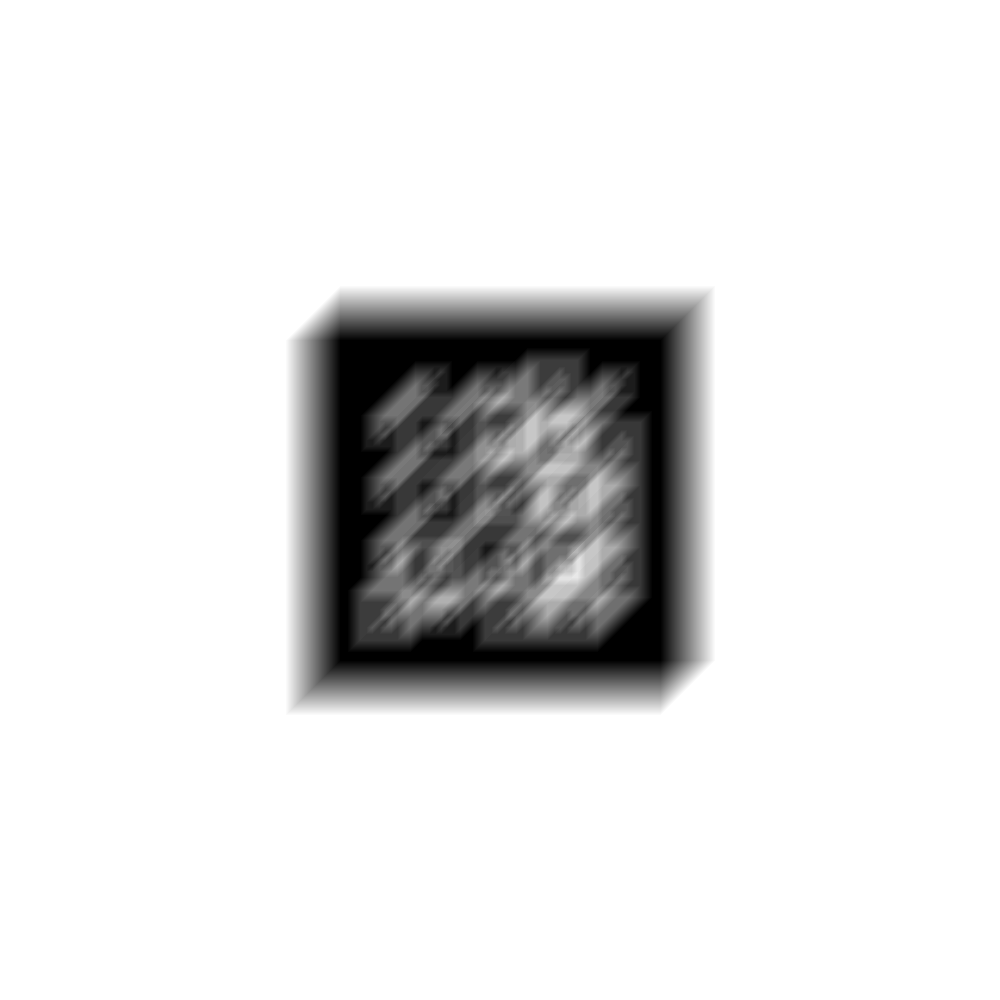}
  \caption*{RArUco (15\% blur)}
\end{minipage}
\caption{Examples of markers subjected to simulated linear motion blur with a kernel size equal to 15\% of each marker's width.}
\label{fig:blur_samples}
\end{figure}

\textbf{Experimental Setup:} All three marker types were generated with an equivalent spatial footprint. We then applied a normalized linear motion blur filter to the images, varying the blur kernel size from 2\% up to 40\% of the marker's physical width. For each blur level, we averaged the detection rates across 100 independent random blur angles (between $0^\circ$ and $180^\circ$). Figure~\ref{fig:blur_samples} shows the visual degradation at a 15\% motion-blur level.

\begin{figure}[htbp]
\centering
\includegraphics[width=0.8\linewidth]{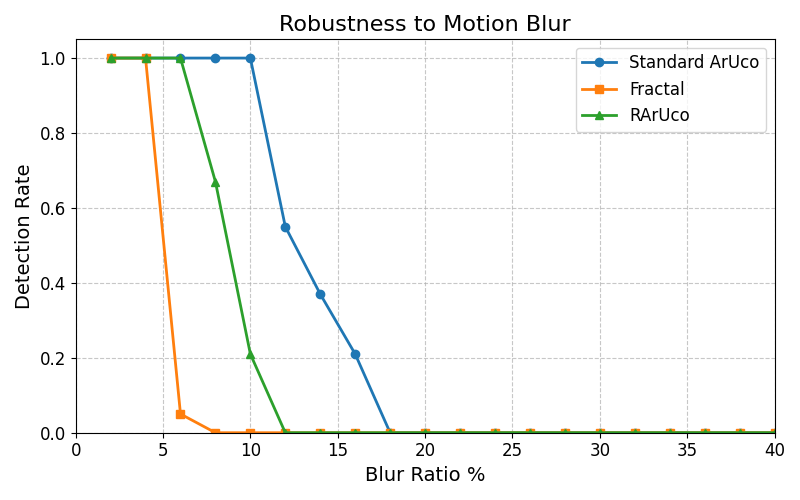}
\caption{Detection rates for standard ArUco, Fractal, and RArUco markers as a function of motion-blur severity. RArUco performs favorably compared to Fractal markers, demonstrating blur resilience nearly on par with standard non-recursive markers.}
\label{fig:robustness_blur}
\end{figure}

\textbf{Results:} Figure~\ref{fig:robustness_blur} plots the detection rates under increasing blur severity. As expected, standard ArUco markers are the most resilient to blur, sustaining 100\% detection up to a 10\% blur ratio, since their bits are physically larger and lack internal recursive subdivisions. The Fractal marker's detection drops off rapidly after 4\% blur, as the high-frequency details of its central nested markers are easily washed out. Crucially, RArUco demonstrates a resilience that closely rivals the standard ArUco marker, maintaining 100\% robust detection up to a 6\% blur ratio and continuing partial detection up to 10\%. This confirms that relying on border bits for parent marker detection avoids the severe fragility associated with purely central sub-marker dependencies, making RArUco highly viable for dynamic UAV environments.

\subsection{Real-World Flight Test}
To complement our simulated benchmarks, we performed a real-world flight test to validate the RArUco marker under actual flight dynamics, lighting variations, and sensor noise. 

\textbf{Experimental Setup:} A DJI Mini 4 Pro drone was flown outdoors over a physical $30 \times 30$ cm RArUco landing pad. The flight trajectory consisted of a vertical takeoff, ascending to an approximate peak altitude of 15 meters, and a subsequent descent to land back on the pad. The onboard camera recorded the maneuver at 30 frames per second (fps), yielding a continuous sequence of approximately 900 frames. The resulting video feed, alongside telemetry data, was collected for analysis.

\begin{figure}[htbp]
\centering
\includegraphics[width=1\linewidth]{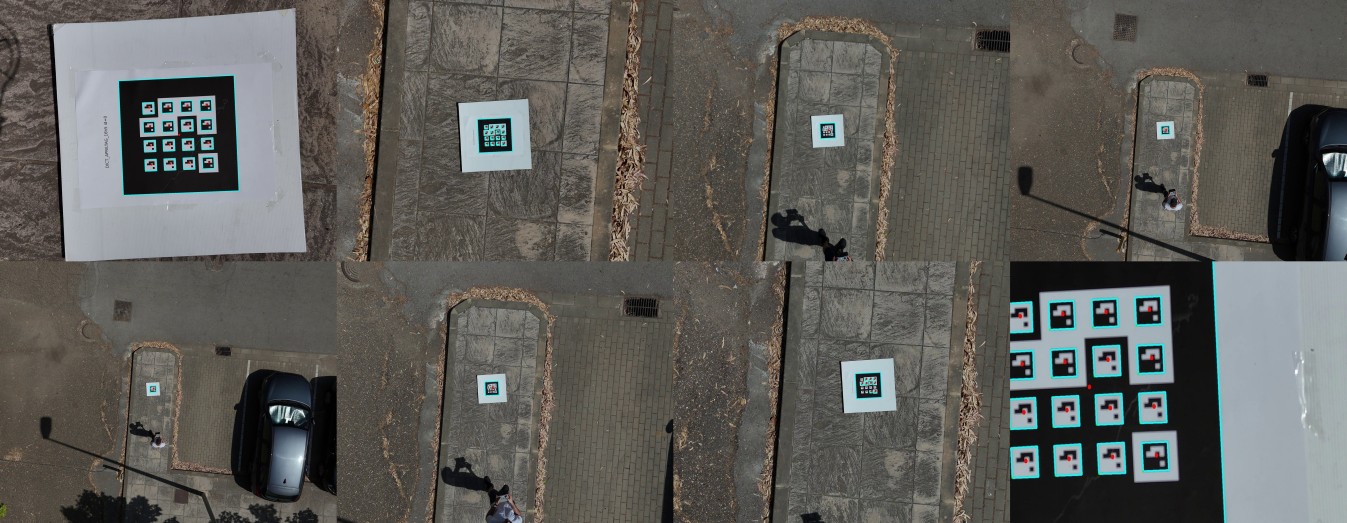}
\caption{Takeoff and landing example. Video footage from a DJI Mini 4 Pro while taking off and landing on the proposed landing pad. The physical landing pad was $30 \times 30$ cm.}
\label{fig:real_world_test}
\end{figure}

\begin{figure}[htbp]
\centering
\includegraphics[width=0.9\linewidth]{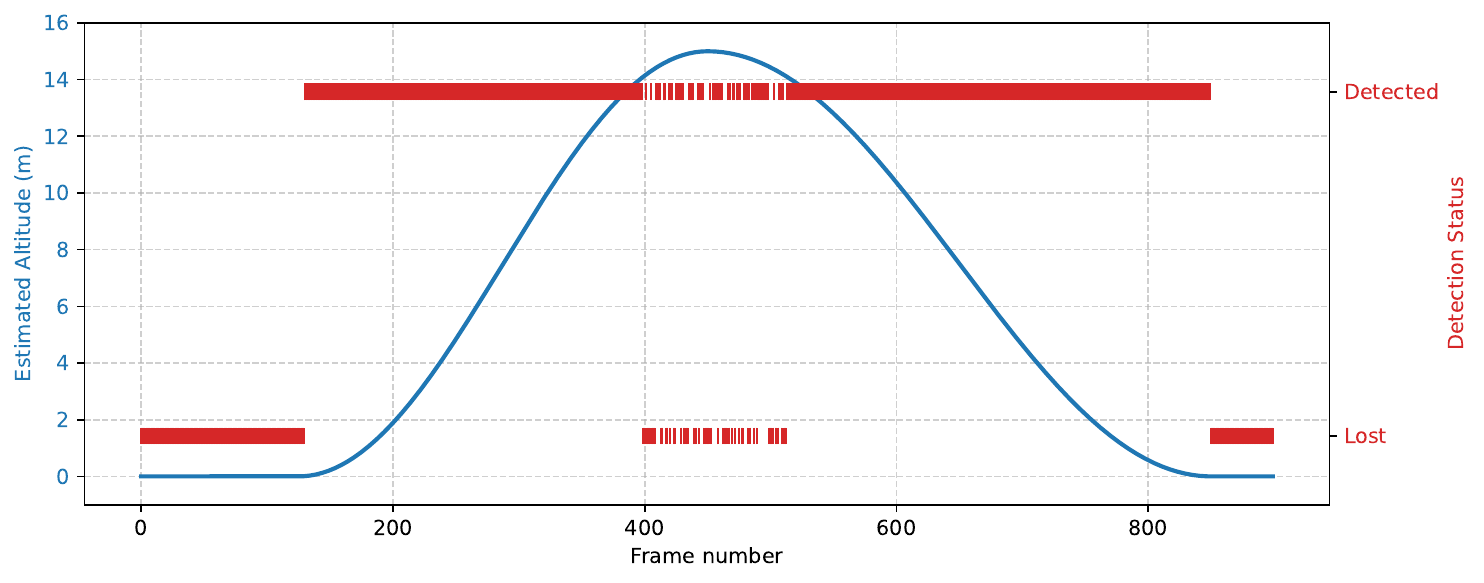}
\caption{UAV altitude and marker detection status across the 900-frame sequence. The proposed marker maintains continuous detection across varying altitudes, exhibiting only the expected instability near the peak altitude of 15 meters, where the marker's apparent size falls below the sensor's decoding resolution.}
\label{fig:flight_plot}
\end{figure}

\textbf{Results:} Figure~\ref{fig:real_world_test} shows representative frames from the video footage during takeoff and landing, while Figure~\ref{fig:flight_plot} shows the estimated altitude and marker detection status throughout the entire flight sequence. The UAV remained stationary on the landing pad for the initial 125 frames, during which the camera was too close to the marker to perceive its internal structure, resulting in no detection. Upon takeoff, the marker was immediately acquired at frame 130. 

The RArUco marker provided robust and continuous tracking throughout the ascent phase. At approximately frame 450, the UAV reached its zenith altitude of 15 meters. Observations indicate that tracking becomes unstable at altitudes exceeding 14 meters, characterized by intermittent detection dropouts. This instability is a natural consequence of the marker's projection falling below the camera sensor's minimum pixel resolution required for robust decoding. As the UAV descended, continuous and stable tracking was re-established. Finally, at frame 850, near the conclusion of the descent, the onboard camera automatically pitched $90^\circ$ forward to prevent mechanical damage upon touchdown, effectively removing the landing pad from the camera's field of view and terminating detection. Overall, the recursive structure permitted the parent marker and its inner sub-markers to synergistically maintain tracking throughout the critical phases of the flight envelope.

\section{Conclusions}
\label{sec::conclusions}

In this paper, we introduce the Recursive ArUco (RArUco) marker, a novel, highly scalable fiducial marker design engineered to overcome the distance limitations inherent in autonomous Unmanned Aerial Vehicle (UAV) landing applications. Unlike traditional recursive approaches that strictly depend on the visibility of the marker's center or severely restrict recursion depth, our proposed method recursively embeds complete markers within both the black and white bits of the parent marker. This structural innovation ensures that detection remains robust even when the marker is heavily occluded or only partially visible within the camera's field of view.

Our specialized detection algorithm, which selectively samples the border regions of each bit, reduces the complexity of the embedded sub-markers. This guarantees reliable bit extraction at any recursion depth without imposing computational overhead. In fact, the proposed method is highly computationally efficient, proving to be faster than the tested existing standard and multi-scale approaches. This superior processing speed renders RArUco particularly well-suited for real-time applications and deployment on resource-constrained embedded devices characteristic of UAV systems. Extensive simulated benchmarks demonstrated that RArUco markers substantially outperform state-of-the-art multi-scale markers, such as Fractal markers, under challenging conditions including steep viewing angles, severe partial occlusion (maintaining a 100\% detection rate up to 30\% random occlusion and 60\% area cropping), and dynamic motion blur.

Furthermore, a real-world flight test using a commercial UAV validated the practical applicability of the RArUco system, confirming its ability to provide continuous, uninterrupted tracking throughout a full 15-meter vertical ascent-and-descent maneuver. By offering arbitrary recursion depth and employing the same dictionary ID across all scales, the RArUco marker presents a mathematically simple, yet highly robust solution for deploying unique and reliable UAV landing pads in complex operational environments. Crucially, preserving a single unique identifier at all scales provides a scalable dictionary of multiple unique landing pads. This allows a fleet of UAVs to operate simultaneously, with each drone navigating to its own designated place---a significant advantage over Fractal and Harco markers.

\bibliographystyle{unsrt}
\bibliography{bibliography}

\end{document}